\documentclass{article}

\usepackage[english]{babel}

\usepackage[letterpaper,top=2cm,bottom=2cm,left=3cm,right=3cm,marginparwidth=1.75cm]{geometry}
\usepackage{booktabs}
\usepackage{amsmath}
\usepackage{graphicx}
\usepackage[pagebackref,breaklinks,colorlinks=true, allcolors=blue]{hyperref}
\usepackage{makecell}
\usepackage[table]{xcolor}
\definecolor{mygray}{gray}{0.9}
\usepackage{mdframed}
\title{Distilling Counterfactual Reasoning from Language to Vision: Causal Graph Guided Post-Training for Video Understanding}
\author{
Yuefei Chen$^{1}$, Jiang Liu$^{2}$, Xiaodong Lin$^{1}$, Ruixiang Tang$^{1\dagger}$ \\[0.5em]
$^{1}$Rutgers University, $^{2}$Advanced Micro Devices \\[0.25em]
{\tt\small chen.yuefei@rutgers.edu, ruixiang.tang@rutgers.edu}
}

\begin{document}

\maketitle
\renewcommand\thefootnote{}\footnotetext{$\dagger$ Corresponding author.}
\begin{abstract}
Vision Language Models (VLMs) have recently shown significant advancements in video understanding, especially in feature alignment, event reasoning, and instruction-following tasks. However, their capability for counterfactual reasoning, inferring alternative outcomes under hypothetical conditions, remains underexplored. This capability is essential for robust video understanding, as it requires identifying underlying causal structures and reasoning about unobserved possibilities, rather than merely recognizing observed patterns.
To systematically evaluate this capability, we introduce CounterVQA, a video-based benchmark featuring three progressive difficulty levels that assess different aspects of counterfactual reasoning. Through comprehensive evaluation of both state-of-the-art open-source and closed-source models, we uncover a substantial performance gap: while these models achieve reasonable accuracy on simple counterfactual questions, performance degrades significantly on complex multi-hop causal chains. To address these limitations, we develop a post-training method, \textbf{CFGPT}, that enhances a model’s visual counterfactual reasoning ability by distilling its counterfactual reasoning capability from the language modality, yielding consistent improvements across all CounterVQA difficulty levels. Dataset and code will be further released. 
\end{abstract}

\section{Introduction}
Visual language models (VLMs) have recently shown remarkable progress in the field of video understanding, especially in tasks such as temporal feature alignment, complex event reasoning, and instruction following \cite{zhang2023video, maaz2024video, wang2024qwen2, bai2025qwen2, yang2025qwen3}. These achievements create an impression that VLMs are rapidly approaching deep cognition and high-level reasoning capabilities in dynamic worlds. However, a deeper sign of intelligence is not only recognizing and describing events that have occurred, but understanding why events occured, then reasoning alternative outcomes under different interventions. This ability, that is also called counterfactual reasoning, is to envision alternate outcomes based on different choices or events, represents the pinnacle of Pearl's Causal Hierarchy \cite{pearl2018book} and underpins core human cognitive capacities including decision-making, moral reasoning, and problem-solving. Reaching counterfactual reasoning ability is widely considered a prerequisite for building robust, generalizable AI and moving towards Artificial General Intelligence \cite{lake2017building, marcus2020next}.

Despite recent progress, several studies suggest that current VLMs may not truly reason about causality.
Work on temporal reasoning and hypothetical queries has shown that VLMs often rely on superficial correlations or linguistic shortcuts rather than understanding underlying causal relations \cite{li2025videohallu, brooks2024video, zhang2025cf}.
These findings imply that VLMs may also struggle with counterfactual reasoning, however, no existing research has systematically evaluated this ability in video settings or investigated how to improve it. Motivated by this gap, we aim to address two key questions:
\begin{figure*}[htbp]
  \centering
  \includegraphics[width=1\textwidth]{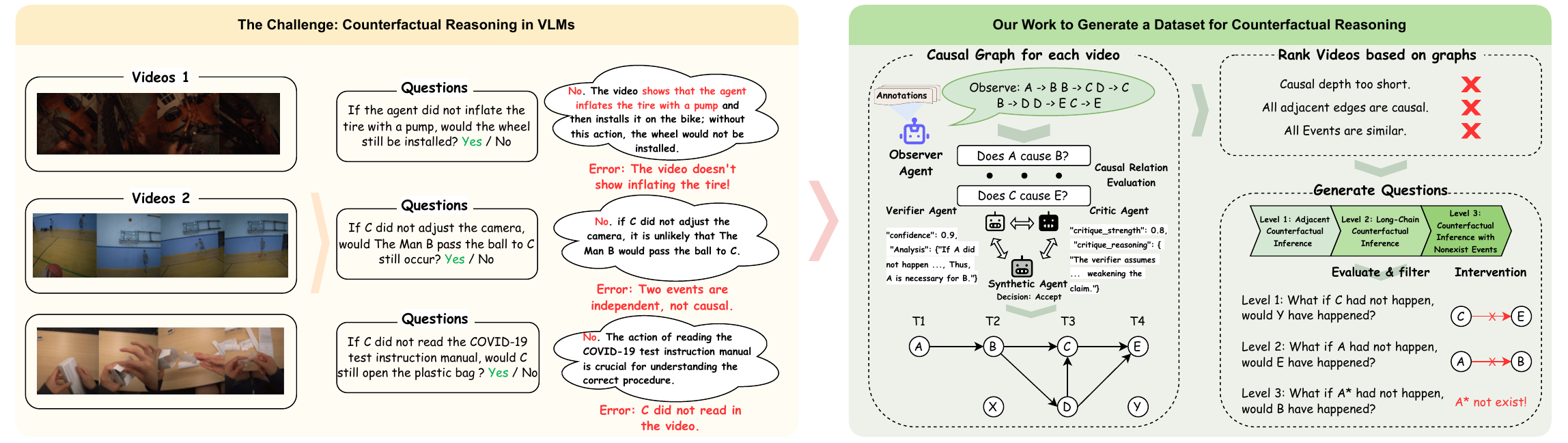}
  \caption{The left panel illustrates typical wrong cases where current VLMs misinterpret causal relations when answering counterfactual video questions. The right panel presents our dataset generation pipeline: a multi-agent system infers pairwise causal relations to build causal graphs for each video, ranks videos by causal graph complexity, and generates three levels of counterfactual questions, adjacent, long-chain, and non-existent event inference, for systematic evaluation.}
  \label{fig:introduction}
\end{figure*}
\paragraph{How well do VLMs handle counterfactual video reasoning?}
Recent benchmarks have advanced the evaluation of causal reasoning in VLMs along several dimensions. Some works \cite{chen2024cello, komanduri2025causalvlbench, wang2025timecausality} evaluate causal structure inference in static images. Other works \cite{hua2024mmcomposition, zhang2024vinoground,thrush2022winoground} assess temporal and compositional reasoning through language. The other works \cite{li2025videohallu, du2024uncovering, liu2025causal3d} probe physical reasoning and anomaly detection in videos. 

Our work complements these efforts by introducing a benchmark CounterVQA, the first video benchmark that systematically generates and verifies counterfactual questions grounded in explicit causal graphs.
Unlike previous datasets, CounterVQA employs a multi-agent generation framework that simulates human-like causal reasoning. This graph-based and agent-driven design allows precise control over causal chain complexity, enabling the construction of three progressive difficulty levels questions that rigorously test reasoning from single-step to complex multi-hop causal chains. 


\paragraph{How to enhance VLM's counterfactual video reasoning abilities?}
Simply scaling up supervision or performing conventional fine-tuning is insufficient, as it tends to reinforce overfitting to observed patterns rather than fostering true causal understanding. What is needed is a causally informed training paradigm that can teach models to align video perception with intervention-based reasoning. 

To address this challenge, we propose a novel post-training framework that integrates supervised fine-tuning (SFT) with reinforcement learning (RL) training \cite{dao2025alphamaze, chen2025sft, guo2025deepseek} guided by causal graph rewards. The central innovation lies in introducing a Causal Graph Reward, a structured signal derived from the explicit causal graphs used in CounterVQA’s multi-agent generation process. During training, the language modality acts as a teacher that excels at logical abstraction, while the visual modality learns as a student to perceive and internalize these abstract causal relations from video evidence. The causal graph reward quantifies consistency between predicted and ground-truth causal structures and it encourages the model to move beyond pattern recognition and to reason over the underlying causal mechanisms that govern event transitions. Empirical results demonstrate that this causally informed post-training improves performance across all difficulty levels in CounterVQA. In summary, our main contributions are as follows. 
\begin{itemize}
    \item We introduce \textbf{CounterVQA}, one of the largest benchmark for assessing counterfactual reasoning ability in VLMs for various situations, designed to evaluate short-chain causal relation, long-chain counterfactual reasoning, and hallucination in counterfactual reasoning. 
    \item We propose a two-stage post-training methodology \textbf{CFGPT} guided by counterfactual causal graphs, which improves VLMs' ability to perform causal and counterfactual reasoning over video content.
    \item We benchmark CounterVQA using five state-of-the-art open- and closed-source models, and demonstrate the effectiveness of CFGPT, which achieves a 12.5\% improvement over the leading 8B-scale VLM across all categories and difficulty levels.
    
\end{itemize}

\section{CounterVQA: Benchmark for VLM video counterfactual reasoning}

To evaluate counterfactual reasoning in VLMs, we introduce CounterVQA, a video-based benchmark with 700+ videos and 3,000+ QA pairs. The benchmark features a two-tier taxonomy:  (1) three levels of counterfactual complexity, and (2) interaction type (Human-to-Human vs. Human-to-Object). 

\subsection{Counterfactual Challenge Levels}
Counterfactual reasoning requires models to imagine alternative outcomes under hypothetical conditions that deviate from observed reality. To systematically assess this capability, we design a three-level hierarchy as shown in Figure~\ref{fig:multi}. This hierarchy progressively increases reasoning complexity. Each level targets distinct aspects of causal understanding: from immediate causal effects to long-range temporal dependencies and resistance to plausible distractors. This hierarchical design allows us to precisely characterize current VLMs' performance in counterfactual reasoning.

\subsubsection{Level 1: Adjacent Counterfactual Inference}

Adjacent counterfactual inference evaluates a model's ability to reason about how intervening on one action affects its immediate successor in an event chain. This represents the most fundamental form of counterfactual reasoning, requiring two distinct capabilities. First, models must differentiate causal relationships from temporal correlations, since adjacent events may co-occur without causal dependency. Second, given a validated causal link, models must predict how an intervention on the cause propagates to the effect. By testing adjacent inference, we assess whether VLMs possess the foundational ability to identify and reason over local causal dependencies, which is prerequisite for more complex multi-hop reasoning tasks.

\paragraph{Formulation.} 
Let $\mathcal{A} = \{a_1, a_2, \ldots, a_n\}$ denote a sequence of actions observed 
in the video, where $a_i$ causally precedes $a_{i+1}$. An adjacent counterfactual 
question intervenes on action $a_i$ with an alternative $a'_i$ and asks about the 
resulting change to the immediate next action:
\begin{equation}
    \text{Query: } a_i \rightarrow a'_i \implies a_{i+1} \rightarrow \, ?
\end{equation}
The model must predict the modified outcome $a'_{i+1}$ under the counterfactual condition $a_i \rightarrow a'_i$. This requires both identifying the causal nature of the relationship $a_i \rightarrow a_{i+1}$ distinguishing it from mere temporal adjacency and reasoning about how the intervention propagates through this causal link while holding other factors constant. 
\paragraph{Example.}
In Figure~\ref{fig:multi} coffee-making video, We assume the original action $a_i$ is "pour milk into the cup" leading to $a_{i+1}$ = "add hot water into the coffee mug", the counterfactual question asks: "If pouring milk had not happened ($\neg a_i$), would the agent still add hot water into the coffee mug?" The model must recognize that these are adjacent but independent events. That is pouring milk does not causally affect adding hot water. 

\subsubsection{Level 2: Long-Chain Counterfactual Inference}

Long-chain counterfactual inference examines whether models can trace the cascading 
effects of an intervention across multiple steps in a causal sequence. Unlike adjacent inference, this level requires reasoning about non-adjacent actions where the intervention propagates through intermediate events before affecting the target outcome. This tests VLMs' capacity for multi-hop causal reasoning and their ability to maintain counterfactual consistency over extended temporal horizons. These capabilities are essential for understanding complex real-world scenarios where actions have delayed or indirect consequences.
\begin{figure}[htbp]
  \centering
  \includegraphics[width=0.6\textwidth]{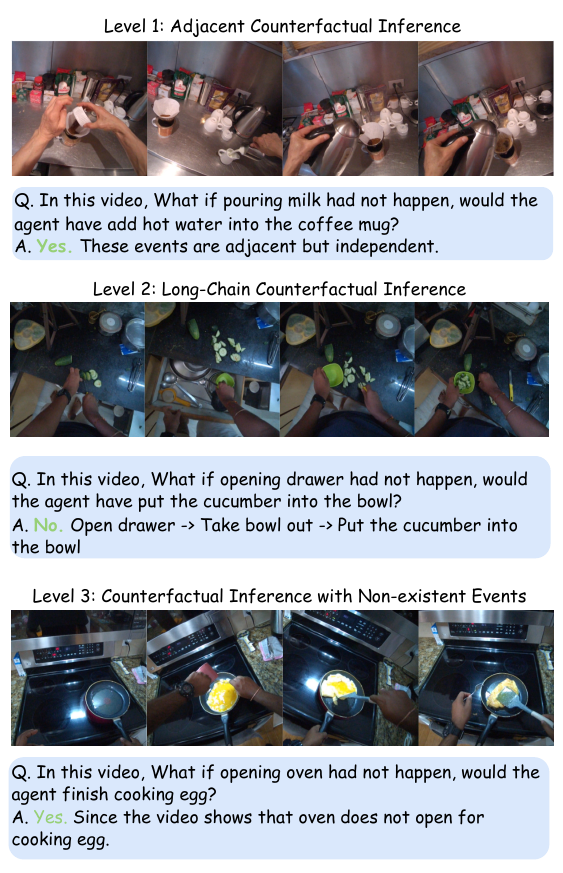}
  \caption{Representative examples from the three difficulty levels of CounterVQA. \textbf{Level 1:} Adjacent counterfactual inference requires reasoning about direct causal dependencies between consecutive events. \textbf{Level 2:} Long-chain counterfactual inference involves tracing multi-hop causal relationships across several actions. \textbf{Level 3:} Counterfactual inference with non-existent events demands reasoning about hypothetical scenarios that did not occur in the observed video.}
  \label{fig:multi}
\end{figure}
\paragraph{Formulation.}
Given an action sequence $\mathcal{A} = \{a_1, \ldots, a_n\}$ and a causal chain 
$a_i \rightarrow a_{i+1} \rightarrow \cdots \rightarrow a_{i+k}$ where $k \geq 2$, 
a long-chain counterfactual question intervenes on $a_i$ and queries about the effect on a non-adjacent action $a_{i+k}$:
\begin{equation}
    \text{Query: } a_i \rightarrow a'_i \implies a_{i+k} \rightarrow \, ?
\end{equation}
The model must reason through the intermediate causal links $a'_i \rightarrow a'_{i+1} \rightarrow \cdots \rightarrow a'_{i+k}$ to predict the final outcome. This requires tracking how changes propagate through the causal graph over multiple hops.
\paragraph{Example.}
In Figure~\ref{fig:multi} cooking preparation video, we assume $a_i$ = "open drawer", which leads through $a_{i+1}$ = "take bowl out" to $a_{i+2}$ = "put the cucumber into the bowl", the counterfactual asks: "If opening drawer had not happened ($\neg a_i$), would the agent have put the cucumber into the bowl ($a_{i+2}$)?" The model must trace the multi-step causal chain: opening drawer $\rightarrow$ taking bowl out $\rightarrow$ putting cucumber. Without the initial action, the entire chain collapses.

\subsubsection{Level 3: Counterfactual Inference with Non-existent Events}
This level evaluates the model's ability to reason about hypothetical events that never occurred in the observed video. A critical challenge in counterfactual reasoning is distinguishing between what actually happened and what could have happened but didn't. VLMs must not only perform counterfactual inference but also verify whether the premise of the counterfactual question is grounded in actual visual observations. This level tests whether models can correctly identify and reason about actions that are semantically plausible given the video context but factually absent from the visual evidence. By requiring models to verify event occurrence before counterfactual inference, we assess their ability to maintain factual grounding while engaging in hypothetical reasoning. It is a critical capability for reliable deployment in real-world applications where distinguishing observed facts from plausible assumptions is essential.

\paragraph{Formulation.}
Let $\mathcal{A}_{\text{true}} = \{a_1, \ldots, a_n\}$ denote the ground-truth action 
sequence and $\mathcal{A}_{\text{distractor}} = \{\tilde{a}_1, \ldots, \tilde{a}_m\}$ 
denote a set of plausible but non-existent actions that are semantically coherent with 
the video context. A Level 3 question presents a counterfactual scenario involving 
$\tilde{a}_j \in \mathcal{A}_{\text{distractor}}$:
\begin{equation}
    \text{Query: } \tilde{a}_j \rightarrow \tilde{a}'_j \implies a_k \rightarrow \,?
\end{equation}
The correct response is to recognize that $\tilde{a}_j$ never occurred, thus rendering the counterfactual premise invalid. This tests the model's ability to distinguish $\mathcal{A}_{\text{true}}$ from $\mathcal{A}_{\text{distractor}}$ under challenging conditions where semantic plausibility conflicts with factual accuracy.

\paragraph{Example.}
In the Figure~\ref{fig:multi} cooking egg video showing only an unopened oven with an egg inside, the question asks: "If opening oven had not happened ($\neg \tilde{a}_j$, where $\tilde{a}_j$ is a non-existent event), would the agent finish cooking egg?" The model must recognize that the oven never actually opened in the video. Since the oven remained closed throughout, no cooking occurred. 
\subsection{Interaction Type}
We divide videos into two categories based on agent interaction type, as each requires distinct causal reasoning skills.
\textbf{Human-to-Human (H2H)} interactions involve collaborative activities such as basketball, soccer, or dance, where reasoning must capture social dynamics and mutual dependencies between agents.
\textbf{Human-to-Object (H2O)} interactions focus on tasks where a person manipulates objects, such as cooking, climbing, or COVID-19 testing, which requires understanding of physical causality and how actions affect material states.
By contrasting H2H and H2O, we assess whether VLMs comprehend both social and physical causality, revealing whether their reasoning generalizes across causal domains.
\subsection{Dataset Generation}
\label{sec:2.3}
\subsubsection{Dataset Collection}
We build our benchmark on Ego-Exo4D~\cite{grauman2024ego}, a large-scale multi-view video dataset comprising 5,035 videos that capture skilled human activities from synchronized egocentric and exocentric perspectives. The dataset covers diverse scenarios including physical activities such as soccer, basketball, and dance, as well as procedural tasks such as cooking and bike repair. To ensure high-quality counterfactual reasoning evaluation, we develop a three-stage pipeline to select suitable videos and generate challenging questions from random 1200 videos. This process filters the original collections down to around 700 and produces over 3,000 question-answer pairs, as detailed below.

\subsubsection{Counterfactual Questions Generations}
Generating high-quality counterfactual questions for videos is challenging: naive LLM-based generation often produces questions that are either too simple or lack grounding in the video's causal structure. To address this, we propose a three-stage pipeline that progressively refines question quality. First, we construct explicit causal graphs using multi-agent systems to capture the key event dependencies in each video. Second, we design graph-based complexity metrics to filter videos that are suitable for counterfactual reasoning. Finally, we employ LLMs to generate questions and use paired LLM verification to ensure answer quality. This approach increases question difficulty substantially: models' accuracy drops by 12\% compared to questions generated by baseline LLM-only methods, validating the effectiveness of our pipeline.
\paragraph{Step1: Multi-Agent Causal Graph Construction.}
For each video, we employ a multi-agent system to construct a causal graph G = (V, E), where nodes V represent key events and edges E capture causal dependencies. Specifically, we use 4 specialized agents. 
\textbf{Observer Agent}: We first generate a set of candidate causal relations and let the LLM assign each relation a confidence score, representing the model’s self-estimated likelihood that the relation is correct. We then rank all candidate relations by these confidence scores and retain only the top 20\% for further evaluation by the Verifier Agent and the Critique Agent.
\textbf{Verifier Agent}: Evaluate a single proposed causal link between two actions based on Commonsense Knowledge.
\textbf{Critic Agent}: The goal is to find flaws in the Verifier's causal analysis. 
\textbf{Synthesizer Agent}: Make a final decision for this causal relational decision based on confidence scores from Verifier and Critic Agents. After that, all accepted causal relations are aggregated into a causal graph for each video. Human evaluation indicates that roughly 90\% of the extracted relations align with human judgments of causal correctness, as detailed in Appendix D.

\paragraph{Step2: Video Filtering via Graph Complexity Metrics.}

Not all videos are equally suitable for counterfactual reasoning. To ensure our benchmark contains videos with rich causal structures across all three difficulty levels, we design three graph-based metrics that directly correspond to the capabilities tested at each level. These metrics quantify different aspects of causal complexity and enable systematic filtering of suitable videos. \textbf{Adjacent Non-Causal Density (ANCD)}: To assess whether a video contains sufficient temporal correlations that must be distinguished from true causality, we compute the ratio of adjacent non-causal relationships to all adjacent relationships:
\begin{equation}
\text{ANCD}(G) = \frac{\#\text{ Adjacent Non-causal Relations}}{\#\text{ Adjacent Relations}}
\end{equation}
A higher value indicates more challenging scenarios for Level 1, as models must 
carefully differentiate spurious temporal adjacency from genuine causal links. 
We retain videos with ANCD $\geq 0.2$ to ensure sufficient 
difficulty in identifying true causal dependencies.
\textbf{Causal Depth (CD)}: To evaluate a video's suitability for long-chain counterfactual reasoning, we measure the maximum path length in the causal graph:
\begin{equation}
    \text{Causal-Depth}(G) = \max_{p \in P(G)} \text{length}(p)
\end{equation}
where $P(G)$ denotes the set of all simple paths in the causal graph $G$, i.e., paths that do not revisit any node. Videos with greater causal depth contain longer causal chains, enabling multi-hop reasoning questions. We select videos with Causal-Depth $\geq 3$ to ensure adequate complexity for Level 2 questions. \textbf{Contextual CNDA Score}: 
Level 3 questions require plausible but non-occurring actions as distractors. We compute the Contextual CNDA score to quantify semantic diversity within each video: videos with diverse actions provide richer semantic contexts where additional plausible alternatives can be naturally conceived. For each action node $v$ in the graph, we first calculate its outlier score based on semantic deviation from its task context $\tau(v)$:
\begin{equation}
    \mathbf{c}_{\tau(v)} = \frac{1}{|C_{\tau(v)}|} \sum_{u \in C_{\tau(v)}} e(u)
\end{equation}
where $C_{\tau(v)}$ is the set of actions sharing the same task context, and 
$e(u)$ is the embedding of action $u$. The outlier score measures how 
semantically atypical an action is:
\begin{equation}
    \text{OutlierScore}(v) = 1 - \frac{e(v) \cdot \mathbf{c}_{\tau(v)}}{\|e(v)\| \cdot \|\mathbf{c}_{\tau(v)}\|}
\end{equation}
We then average across all nodes to obtain the video-level CNDA score:
\begin{equation}
    \text{Avg-CNDA-Score}(G) = \frac{1}{|V|} \sum_{v \in V} \text{OutlierScore}(v)
\end{equation}
A higher CNDA score indicates that the video contains diverse actions, making it 
easier to construct plausible hallucinated alternatives. We retain videos with 
avg-CNDA-score $\geq 0.12$ to ensure sufficient semantic diversity for Level 3 
distractor generation. These metrics filtering reduces the candidate set from 1,200 videos to 700+, ensuring that retained videos exhibit rich causal structures necessary for challenging counterfactual reasoning evaluation.

\paragraph{Step3: Question Generation and Quality Control.}
For each filtered video, we use Deepseek-V3 to generate counterfactual questions based on the causal graph. The generation prompt instructs the LLM to identify a pivotal node in the graph, hypothesize an alternative outcome at that node, and  reason about downstream effects. To ensure question quality, we implement a paired verification process. Two independent text-only LLMs (Qwen 2.5-32B) attempt to answer each generated question using only the video captions. We only keep the challenging questions which can be answered correctly by LLMs with video annotations. This verification step removes approximately 80\% of generated questions, resulting in a final set of 3,000+ high-quality QA pairs. To ensure the reliability of causal graph and questions, we are conducting a validation on the dataset, including inter-annotator agreement and relation-type statistics. The detailed annotation consistency results and question-type distributions are included in the Appendix C.

\section{Counterfactual Graph-based Post-Training (CFGPT)}
Beyond benchmarking, we explore how to systematically enhance VLMs' counterfactual reasoning in videos. Our investigation reveals two key insights into current VLMs' limitations: \textbf{1. Cross-modal reasoning gap}: While VLMs demonstrate strong causal reasoning in text, they cannot ground these capabilities in video understanding, suggesting difficulties in transferring reasoning patterns from linguistic to visual-temporal domains \cite{yu2024mm,parcalabescu2022valse}. \textbf{2. Lack of structural guidance}: VLMs frequently produce counterfactual reasoning that either invents visual content or creates inconsistent causal chains, indicating insufficient supervision from both visual evidence and explicit causal structures.
\begin{table}[htbp]
\centering
\scalebox{0.72}{
\begin{tabular}{lcc}
\toprule
\textbf{Teacher Setting} & \textbf{Accuracy (\%)} & \textbf{Readability Score (5.0)} \\
\midrule
Video   &  41.9  &  3.2 \\
Video+Annotations &  \textbf{46.7}  &  \textbf{3.5} \\
\bottomrule
\end{tabular}}
\caption{Comparison of outputs under video+annotation (V+A) and video-only (V) prompting. Readability metric details are in Appendix~E.}
\label{table:teacher_quality}

\end{table}
To probe the source of these limitations, we test whether reasoning improves when explicit temporal and relational cues are provided. We compare two prompting settings on Qwen-2.5-VL 32B: using only raw video versus using video with structured annotations (V+A). This allows us to assess whether adding causal structure yields more coherent counterfactual reasoning. As shown in Table~\ref{table:teacher_quality}, V+A consistently produces higher accuracy and more logically grounded explanations. These findings indicate that structured causal cues substantially enhance reasoning quality and motivate a training paradigm that can transfer these improvements to video-only models.

Building on this insight, we propose \textbf{Counterfactual Graph-based Post-Training (CFGPT)}, a two-stage framework for strengthening video-grounded counterfactual reasoning (Figure~\ref{fig:introduction}).
Stage I (Cross-Modal Causal Transfer) distills textual causal reasoning into video-grounded inference using structured supervision.
Stage II (Visual-Causal Reinforcement) further refines this skill with rewards from visual grounding and causal-graph alignment, ensuring both factual correctness and causal consistency.
\begin{figure*}[htbp]
  \centering
  \includegraphics[width=0.75\textwidth]{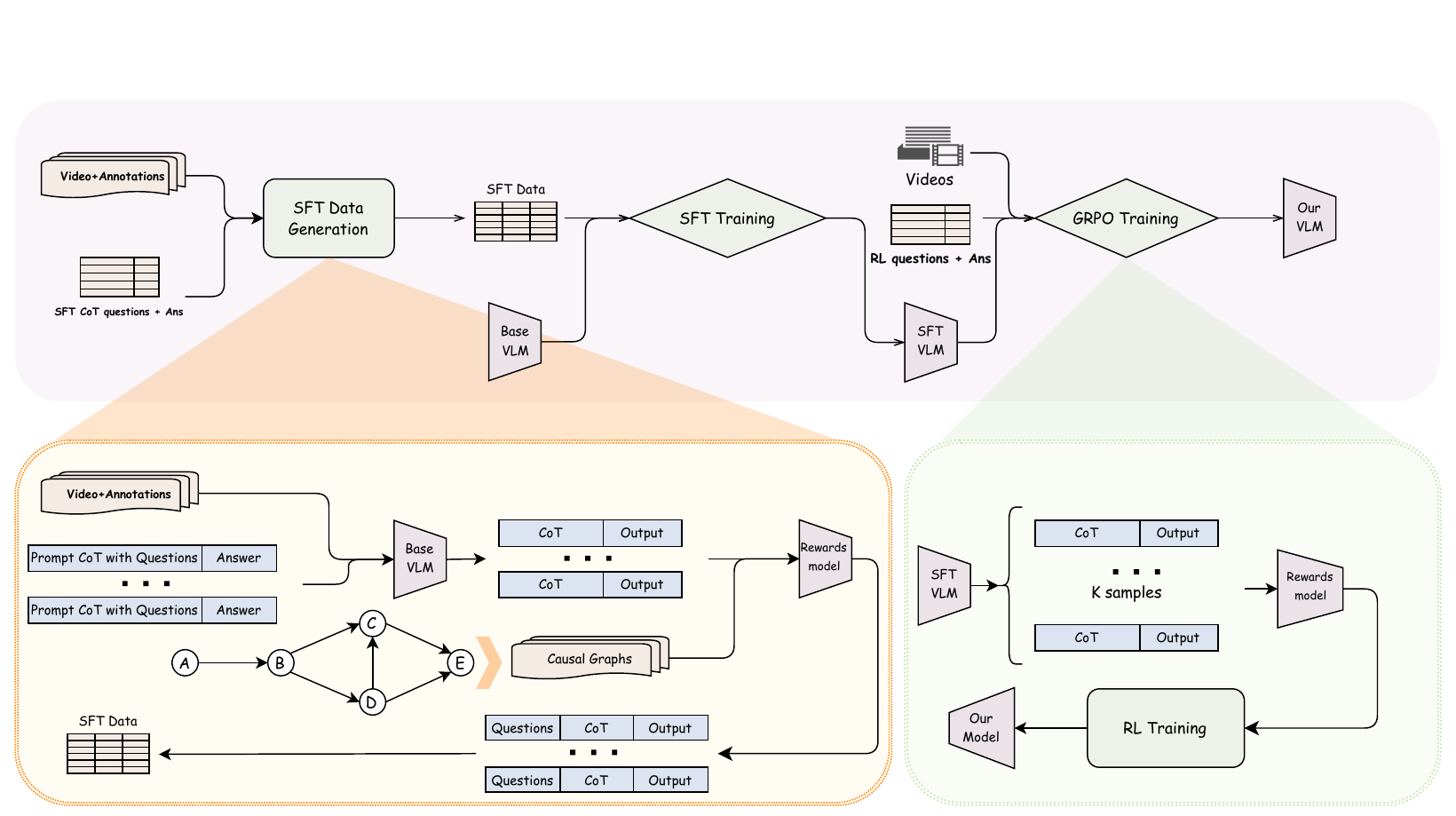}
  \caption{Overview of the CFGPT framework. \textbf{Top:} Two-stage pipeline from a base VLM to the final CFGPT model.
\textbf{Left (orange):} Cross-modal causal transfer through supervised fine-tuning using V+A generated CoT data.
\textbf{Right (green):} Visual-causal reinforcement via GRPO, where candidate outputs are scored by causal graph consistency and visual grounding rewards to refine counterfactual reasoning.}
  \label{fig:method}
\end{figure*}
\subsection{Cross-Modal Causal Transfer Knowledge}
The first stage of CFGPT enables the transfer of causal reasoning capabilities from text-based understanding to video-grounded inference. While language models excel at reasoning about causal relationships described in text, VLMs struggle to perform similar reasoning when grounded in dynamic visual content. We bridge this gap through carefully designed supervised fine-tuning that teaches the model to ground textual causal reasoning patterns in temporal video understanding. To support this process, We construct training data that explicitly connects textual causal reasoning with video content. Starting from videos with rich annotations, we leverage LLMs to generate chain-of-thought rationales that articulate causal relationships observable in the video. This produces training instances in the format of \textit{(video, question, CoT reasoning, answer)}, where the CoT serves as a bridge connecting visual observations to causal inference, enabling cross-modal knowledge transfer. Given the base VLM denoted as $f_{VLM}$, we fine-tune it on the SFT data $\mathcal{D}_{SFT} = \{(v_i,q_i,c_i,a_i)\}^N_{i=1}$, where $v_i$ represents the video, $q_i$ is the question, $c_i$ is the chain-of-thought rationale, and $a_i$ is the answer. The training objective is: 
\begin{equation}
\mathcal{L}_{\text{SFT}} = -\sum_{i=1}^N \left[\log p(c_i | v_i, q_i) + \log p(a_i | v_i, q_i, c_i)\right]
\end{equation}
This objective encourages the model to first generate coherent reasoning chains that capture the causal dependencies, and then produce accurate answers conditioned on both the video context and the reasoning process.
 The explicit CoT rationales serve as the crucial bridge to connect Textual and Visual Reasoning in this transfer process. By verbalizing the reasoning process grounded in visual observations, the model learns to translate its textual causal understanding into video-specific inference patterns. This enables the model to identify temporal dependencies, recognize causal events in motion, and reason about counterfactual scenarios based on what is actually shown in the video.
\subsection{Visual-Causal Alignment Optimization}
While the transfer learning stage establishes basic video-grounded reasoning, the model may still exhibit inconsistencies between its visual understanding and causal inference. To address this, we employ reinforcement learning with dual guidance from both visual content and explicit causal graph structures, ensuring the model's reasoning is simultaneously grounded in what it observes and structurally sound in its causal logic.
For each training video, we construct a counterfactual causal graph $\mathcal{G} = (\mathcal{V}, \mathcal{E})$ that explicitly represents the causal relationships between events and entities mentioned in section 2.3 Multi-agent part. The nodes $\mathcal{V}$ correspond to key events in the video, while edges $\mathcal{E}$ encode the causal dependencies. These graphs serve as structural guidance to evaluate whether the model's reasoning aligns with correct causal pathways.
We design a reward model that assesses the quality of the model's outputs along multiple dimensions:
\begin{equation}
R(o | v, q, \mathcal{G}) = \alpha R_{\text{causal}}(o, \mathcal{G}) + \beta R_{\text{visual}}(o, v)
\end{equation}
where $o$ represents the model's output (including both CoT and answer), $R_{\text{causal}}$ evaluates alignment with the causal graph structure, and $R_{\text{visual}}$ assesses the alignment between the VLM's understanding and the actual video content. The hyperparameters $\alpha, \beta$ balance these objectives.
While the transfer learning stage establishes basic video-grounded reasoning, the model may still exhibit inconsistencies between its visual understanding and causal inference. To address this, we employ reinforcement learning with dual guidance from both visual content and explicit causal graph structures, ensuring the model's reasoning is simultaneously grounded in what it observes and structurally sound in its causal logic. Following the GRPO framework, we sample $K$ outputs for each input and optimize through relative comparison. Through iterative sampling, dual reward evaluation, and policy updates, the model learns to consistently generate reasoning that aligns with both visual observations and causal structures. This dual supervision helps the model avoid two common failure modes: semantic related content not present in videos ($R_{\text{visual}}$) and making logically inconsistent causal inferences ($R_{\text{causal}}$), leading to more reliable video understanding.

\section{Experiment}
\begin{table*}[h!]
\centering
\scalebox{0.61}{
\begin{tabular}{p{5.6cm}|>{\centering\arraybackslash}p{1.3cm}>{\centering\arraybackslash}p{1.3cm}>{\centering\arraybackslash}p{1.3cm}>{\centering\arraybackslash}p{0.8cm}|>{\centering\arraybackslash}p{1.3cm}>{\centering\arraybackslash}p{1.3cm}>{\centering\arraybackslash}p{1.3cm}>{\centering\arraybackslash}p{0.8cm}|>{\centering\arraybackslash}p{1.3cm}>{\centering\arraybackslash}p{1.3cm}>{\centering\arraybackslash}p{1.3cm}>{\centering\arraybackslash}p{0.8cm}}
\toprule
\rowcolor{mygray} \multicolumn{1}{c}{} & \multicolumn{4}{c}{\textbf{Human to Human}} & \multicolumn{4}{c}{\textbf{Human to Object}} & \multicolumn{4}{c}{\textbf{Total}} \\ \hline
\rowcolor{mygray} \textbf{Models} & \textbf{Level 1} & \textbf{Level 2} & \textbf{Level 3} & \textbf{Avg.} & \textbf{Level 1} & \textbf{Level 2} & \textbf{Level 3} & \textbf{Avg.} & \textbf{Level 1} & \textbf{Level 2} & \textbf{Level 3} & \textbf{Avg.} \\ \hline

\multicolumn{13}{l}{\textit{Open-source (Vanilla)}} \\ 
\hline
Qwen-2.5-VL 7B & $13.3$ & $11.5$ & $3.1$ & $9.1$ & $12.1$ & $9.9$ & $2.5$ & $8.1$ & $12.3$ & $10.3$ & $2.7$ & $8.3$ \\
Qwen-2.5-VL 32B & $28.6$ & $21.4$ & $64.3$ & $36.7$ & $33.5$ & $51.5$ & $41.9$ & $39.7$ & $33.0$ & $46.2$ & $44.0$ & $39.4$ \\
Qwen-3-VL 8B  & $45.1$ & $54.5$ & $52.6$ & $49.1$ & $56.7$ & $63.7$ & $66.4$ & $61.9$ & $54.6$ & $62.1$ & $65.3$ & $60.1$ \\
\midrule
\multicolumn{13}{l}{\textit{Closed-source (API)}} \\ 
\hline
ChatGPT-4o & $12.5$ & $15.8$ & $8.3$ & $12.7$ & $9.0$ & $6.5$ & $0.7$ & $3.9$ & $9.6$ & $8.0$ & $1.3$ & $6.2$ \\
Gemini-2.5-Pro & $57.1$ & $25.0$ & $40.0$ & $43.8$ & $39.3$ & $42.4$ & $42.6$ & $41.3$ & $41.1$ & $40.5$ & $42.4$ & $41.5$ \\
\midrule
\multicolumn{13}{l}{\textit{Open-source (Finetuned / Ours)}} \\ 
\hline
Qwen-2.5-VL 7B + SFT & $12.5$ & $15.4$ & $3.6$ & $10.7$ & $12.8$ & $9.9$ & $3.3$ & $8.7$ & $12.7$ & $11.2$ & $3.4$ & $9.1$ \\
Qwen-3-VL 8B + SFT & $76.3$ & $54.5$ & $\mathbf{73.7}$ & $\mathbf{72.0}$ & $65.9$ & $71.4$ & $70.6$ & $68.8$ & $68.9$ & $67.9$ & $71.0$ & $69.3$ \\
\textbf{Qwen-2.5-VL 7B + SFT + RL} & $12.1$ & $15.4$ & $3.5$ & $10.3$ & $12.8$ & $9.9$ & $4.2$ & $9.0$ & $12.7$ & $11.2$ & $4.1$ & $9.3$ \\
\textbf{Qwen-3-VL 8B + SFT + RL} & $\mathbf{77.4}$ & $\mathbf{57.1}$ & $63.2$ & $68.8$ & $\mathbf{70.6}$ & $\mathbf{73.9}$ & $\mathbf{77.4}$ & $\mathbf{73.8}$ & $\mathbf{70.1}$ & $\mathbf{71.6}$ & $\mathbf{76.0}$ & $\mathbf{72.6}$ \\

\bottomrule
\end{tabular}
}
\caption{Performance of various VLMs on CounterVQA across three difficulty levels and two interaction types. Our finetuned models (SFT and SFT+RL) achieve the highest accuracy overall.}
\label{table:model_comparison}
\end{table*}
\subsection{Evaluation Settings and Models}
We conduct experiments on the proposed CounterVQA benchmark to evaluate the counterfactual reasoning capability of VLMs. 
CounterVQA consists of three difficulty levels, Adjacent Counterfactual Inference, Long-Chain Counterfactual Inference, and Counterfactual Inference with Non-existent Events. Videos in these dataset are categorized into H2H and H2O. We evaluate different vanilla open-source models, including Qwen2.5-VL-7B, Qwen2.5-VL-32B \cite{bai2025qwen2}, and Qwen3-VL-8B \cite{yang2025qwen3}, and other closed source LLMs like GPT-4o \cite{openai_gpt4o_2024} and Gemini-2.5 pro \cite{gemini_pro_2024}. The details of parameters and running environments are described in Appendix A.

\subsection{Main Result and Analysis}
Table~\ref{table:model_comparison} presents the performance of all tested VLMs on CounterVQA, showing accuracy metrics as percentages across three difficulty levels and two interaction types (H2H and H2O). The results reveal several critical observations that validate both the effectiveness of our benchmark and the superiority of our proposed CFGPT method.

\paragraph{Our benchmark effectively exposes limitations in current VLMs.} Despite recent advances, most existing VLMs demonstrate significant limitations in counterfactual reasoning. ChatGPT-4o, a leading closed-source model, achieves only 6.2\% average accuracy. Even the larger Qwen-2.5-VL-32B model struggles with an average accuracy of 39.4\%. These results confirm that counterfactual reasoning in videos remains a fundamental challenge that cannot be solved by simply scaling model parameters, validating the necessity and difficulty of our benchmark.
Beyond overall performance, we further analyze how different types of interactions affect model reasoning capabilities. Comparing the two interaction types, we observe the results do not show a consistent advantage for either Human-to-Human or Human-to-Object reasoning across models. Some models perform better on object-centered interactions, while others handle social interactions more effectively. This variability indicates that current VLMs struggle to generalize causal reasoning uniformly across different interaction types.

\paragraph{CFGPT significantly improves counterfactual reasoning capabilities.} Our method demonstrates substantial and consistent improvements across all difficulty levels and interaction types. Qwen-3-VL-8B + SFT alone improves the vanilla model from 60.1\% to 69.3\% (+9.2 points), demonstrating the effectiveness of cross-modal causal transfer learning. More importantly, adding GRPO further boosts performance to 72.6\% (+12.5 points over vanilla), achieving the best results across all models. The improvement is particularly pronounced on harder levels: on Level 3 H2O tasks, our full method achieves 77.4\% compared to 66.4\% for the vanilla model (+11 points), demonstrating that visual-causal dual reinforcement is especially effective for complex counterfactual scenarios. However, the effectiveness of CFGPT is not uniform across all base models. Model capacity influences the effectiveness of CFGPT. CFGPT yields only modest gains on Qwen-2.5-VL-7B (8.3\% $\rightarrow$ 9.3\%) but substantially improves Qwen-3-VL-8B (60.1\% $\rightarrow$ 72.6\%), indicating that a minimum level of video understanding is required for effective counterfactual reasoning transfer. Qwen-3-VL-8B’s higher baseline provides a better foundation for cross-modal transfer and reinforcement stages.Specialized post-training is more important than scale. Qwen-3-VL-8B (60.1\%) already surpasses larger models such as Qwen-2.5-VL-32B (39.4\%) and Gemini-2.5-Pro (41.5\%), highlighting that architecture and training objectives matter more than parameter count. With CFGPT, Qwen-3-VL-8B reaches 72.6\%, outperforming these larger models by over 30 points. It shows that targeted post-training is essential for strengthening counterfactual reasoning in videos, whereas simply scaling model size or relying on general purpose training is insufficient. Additional ablations study and error analysis are provided in Appendix~B.

\section{Related Work}
\noindent \textbf{Vision Language Models.} Early large-scale VLMs connect pretrained vision encoders with LLMs and learn to align modalities through lightweight adapters or instruction tuning. Representative lines include Flamingo, which interleaves image/video tokens with text for few-shot multimodal prompting \cite{alayrac2022flamingo}; PaLI/PaLI-X that scale both the vision and language components jointly \cite{chen2022pali,chen2023pali}; BLIP-2 that freezes the image encoder and LLM while learning a Querying Transformer as a bridge \cite{li2023blip}; and instruction-tuned systems such as LLaVA that distill multimodal instruction-following behaviors \cite{liu2023visual}. Recent open models introduce dynamic-resolution tokenization and unified image–video processing, improving long-video comprehension \cite{wang2024qwen2,bai2025qwen2}.

\noindent \textbf{Counterfactual Reasoning in VLMs.} Counterfactual reasoning probes whether a model can answer “what if” questions beyond mere pattern association. Recent datasets for images (e.g., C-VQA) explicitly inject counterfactual presuppositions and show that state-of-the-art VLMs still struggle on counterfactual queries despite strong perceptual skills \cite{zhang2024if}. Related benchmarks \cite{zellers2019recognition, agrawal2018don} demonstrate that models often fail when causal, hypothetical, or distribution-shift reasoning is required. Contemporary work also explores counterfactual fine-tuning objectives for VLMs to improve semantic discrimination and causal robustness, indicating that exposing models to minimally edited pairs can help disentangle spurious cues \cite{zhang2025cf}. Overall, the gap between visual recognition and robust counterfactual inference remains a key challenge.

\noindent \textbf {Benchmarks for Video Understanding and Counterfactual Reasoning.} A line of video reasoning benchmarks stresses temporal/causal structure. CLEVRER introduces synthetic collision videos with descriptive, explanatory, predictive, and counterfactual questions, catalyzing research on causal video QA \cite{yi2019clevrer}. CATER focuses on compositional actions and temporal reasoning \cite{girdhar2019cater}. NExT-QA moves from description to explanation, grouping questions into causal and temporal categories over real videos \cite{xiao2021next}. For physically grounded causal reasoning, the Physion benchmark evaluates predictions of physical outcomes from videos \cite{bear2021physion}. Very recent video QA benchmarks foreground hypothetical/anticipation/counterfactual queries on real-world scenarios, and consistently report a large human–model gap \cite{foss2025causalvqa,chen2025compositional}. Causal reasoning benchmarks like CLADDER~\cite{jin2023cladder} and CounterBench~\cite{chen2025counterbench} show LLMs’ weaknesses in multi-step and counterfactual inference.

\section{Conclusion}
We introduce CounterVQA, a benchmark designed to evaluate counterfactual reasoning in VLMs for video understanding. It features multi-agent-annotated, causal-graph-intensive QA pairs under Interaction categories to assess VLM's ability to detect counterfactual reasoning. CounterVQA reveals substantial gaps in VLMs’ causal and counterfactual reasoning. To address these, We propose CFGPT, a post-training framework that aligns model reasoning with video-grounded causal structures. Extensive experiments show effectiveness of this method in solving these issues across models. Our future work will focus on expanding a long activity causal video dataset with synthetic videos to train VLMs for improving video understanding and counterfactual reasoning. 

\bibliographystyle{alpha}
\bibliography{ref}

@article{alayrac2022flamingo,
  title={Flamingo: a visual language model for few-shot learning},
  author={Alayrac, Jean-Baptiste and Donahue, Jeff and Luc, Pauline and Miech, Antoine and Barr, Iain and Hasson, Yana and Lenc, Karel and Mensch, Arthur and Millican, Katherine and Reynolds, Malcolm and others},
  journal={Advances in neural information processing systems},
  volume={35},
  pages={23716--23736},
  year={2022}
}

@article{chen2022pali,
  title={Pali: A jointly-scaled multilingual language-image model},
  author={Chen, Xi and Wang, Xiao and Changpinyo, Soravit and Piergiovanni, Anthony J and Padlewski, Piotr and Salz, Daniel and Goodman, Sebastian and Grycner, Adam and Mustafa, Basil and Beyer, Lucas and others},
  journal={arXiv preprint arXiv:2209.06794},
  year={2022}
}

@article{chen2023pali,
  title={Pali-x: On scaling up a multilingual vision and language model},
  author={Chen, Xi and Djolonga, Josip and Padlewski, Piotr and Mustafa, Basil and Changpinyo, Soravit and Wu, Jialin and Ruiz, Carlos Riquelme and Goodman, Sebastian and Wang, Xiao and Tay, Yi and others},
  journal={arXiv preprint arXiv:2305.18565},
  year={2023}
}

@inproceedings{li2023blip,
  title={Blip-2: Bootstrapping language-image pre-training with frozen image encoders and large language models},
  author={Li, Junnan and Li, Dongxu and Savarese, Silvio and Hoi, Steven},
  booktitle={International conference on machine learning},
  pages={19730--19742},
  year={2023},
  organization={PMLR}
}

@article{liu2023visual,
  title={Visual instruction tuning},
  author={Liu, Haotian and Li, Chunyuan and Wu, Qingyang and Lee, Yong Jae},
  journal={Advances in neural information processing systems},
  volume={36},
  pages={34892--34916},
  year={2023}
}

@article{wang2024qwen2,
  title={Qwen2-vl: Enhancing vision-language model's perception of the world at any resolution},
  author={Wang, Peng and Bai, Shuai and Tan, Sinan and Wang, Shijie and Fan, Zhihao and Bai, Jinze and Chen, Keqin and Liu, Xuejing and Wang, Jialin and Ge, Wenbin and others},
  journal={arXiv preprint arXiv:2409.12191},
  year={2024}
}

@article{bai2025qwen2,
  title={Qwen2. 5-vl technical report},
  author={Bai, Shuai and Chen, Keqin and Liu, Xuejing and Wang, Jialin and Ge, Wenbin and Song, Sibo and Dang, Kai and Wang, Peng and Wang, Shijie and Tang, Jun and others},
  journal={arXiv preprint arXiv:2502.13923},
  year={2025}
}

@inproceedings{zhang2024if,
  title={What if the tv was off? examining counterfactual reasoning abilities of multi-modal language models},
  author={Zhang, Letian and Zhai, Xiaotong and Zhao, Zhongkai and Zong, Yongshuo and Wen, Xin and Zhao, Bingchen},
  booktitle={Proceedings of the IEEE/CVF Conference on Computer Vision and Pattern Recognition},
  pages={21853--21862},
  year={2024}
}

@article{zhang2025cf,
  title={CF-VLM: CounterFactual Vision-Language Fine-tuning},
  author={Zhang, Jusheng and Cai, Kaitong and Fan, Yijia and Wang, Jian and Wang, Keze},
  journal={arXiv preprint arXiv:2506.17267},
  year={2025}
}

@article{yi2019clevrer,
  title={Clevrer: Collision events for video representation and reasoning},
  author={Yi, Kexin and Gan, Chuang and Li, Yunzhu and Kohli, Pushmeet and Wu, Jiajun and Torralba, Antonio and Tenenbaum, Joshua B},
  journal={arXiv preprint arXiv:1910.01442},
  year={2019}
}

@article{girdhar2019cater,
  title={Cater: A diagnostic dataset for compositional actions and temporal reasoning},
  author={Girdhar, Rohit and Ramanan, Deva},
  journal={arXiv preprint arXiv:1910.04744},
  year={2019}
}

@inproceedings{xiao2021next,
  title={Next-qa: Next phase of question-answering to explaining temporal actions},
  author={Xiao, Junbin and Shang, Xindi and Yao, Angela and Chua, Tat-Seng},
  booktitle={Proceedings of the IEEE/CVF conference on computer vision and pattern recognition},
  pages={9777--9786},
  year={2021}
}

@article{bear2021physion,
  title={Physion: Evaluating physical prediction from vision in humans and machines},
  author={Bear, Daniel M and Wang, Elias and Mrowca, Damian and Binder, Felix J and Tung, Hsiao-Yu Fish and Pramod, RT and Holdaway, Cameron and Tao, Sirui and Smith, Kevin and Sun, Fan-Yun and others},
  journal={arXiv preprint arXiv:2106.08261},
  year={2021}
}

@article{foss2025causalvqa,
  title={CausalVQA: A Physically Grounded Causal Reasoning Benchmark for Video Models},
  author={Foss, Aaron and Evans, Chloe and Mitts, Sasha and Sinha, Koustuv and Rizvi, Ammar and Kao, Justine T},
  journal={arXiv preprint arXiv:2506.09943},
  year={2025}
}

@article{chen2025compositional,
  title={Compositional physical reasoning of objects and events from videos},
  author={Chen, Zhenfang and Dong, Shilong and Yi, Kexin and Li, Yunzhu and Ding, Mingyu and Torralba, Antonio and Tenenbaum, Joshua B and Gan, Chuang},
  journal={IEEE Transactions on Pattern Analysis and Machine Intelligence},
  year={2025},
  publisher={IEEE}
}

@book{pearl2018book,
  title={The book of why: the new science of cause and effect},
  author={Pearl, Judea and Mackenzie, Dana},
  year={2018},
  publisher={Basic books}
}

@inproceedings{grauman2024ego,
  title={Ego-exo4d: Understanding skilled human activity from first-and third-person perspectives},
  author={Grauman, Kristen and Westbury, Andrew and Torresani, Lorenzo and Kitani, Kris and Malik, Jitendra and Afouras, Triantafyllos and Ashutosh, Kumar and Baiyya, Vijay and Bansal, Siddhant and Boote, Bikram and others},
  booktitle={Proceedings of the IEEE/CVF Conference on Computer Vision and Pattern Recognition},
  pages={19383--19400},
  year={2024}
}

@article{li2025videohallu,
  title={VideoHallu: Evaluating and Mitigating Multi-modal Hallucinations on Synthetic Video Understanding},
  author={Li, Zongxia and Wu, Xiyang and Shi, Guangyao and Qin, Yubin and Du, Hongyang and Zhou, Tianyi and Manocha, Dinesh and Boyd-Graber, Jordan Lee},
  journal={arXiv preprint arXiv:2505.01481},
  year={2025}
}

@article{zhang2024vinoground,
  title={Vinoground: Scrutinizing lmms over dense temporal reasoning with short videos},
  author={Zhang, Jianrui and Cai, Mu and Lee, Yong Jae},
  journal={arXiv preprint arXiv:2410.02763},
  year={2024}
}

@article{chen2024cello,
  title={Cello: Causal evaluation of large vision-language models},
  author={Chen, Meiqi and Peng, Bo and Zhang, Yan and Lu, Chaochao},
  journal={arXiv preprint arXiv:2406.19131},
  year={2024}
}

@article{dao2025alphamaze,
  title={AlphaMaze: Enhancing Large Language Models' Spatial Intelligence via GRPO},
  author={Dao, Alan and Vu, Dinh Bach},
  journal={arXiv preprint arXiv:2502.14669},
  year={2025}
}

@article{chen2025sft,
  title={Sft or rl? an early investigation into training r1-like reasoning large vision-language models},
  author={Chen, Hardy and Tu, Haoqin and Wang, Fali and Liu, Hui and Tang, Xianfeng and Du, Xinya and Zhou, Yuyin and Xie, Cihang},
  journal={arXiv preprint arXiv:2504.11468},
  year={2025}
}

@article{brooks2024video,
  title={Video generation models as world simulators},
  author={Brooks, Tim and Peebles, Bill and Holmes, Connor and DePue, Will and Guo, Yufei and Jing, Li and Schnurr, David and Taylor, Joe and Luhman, Troy and Luhman, Eric and others},
  journal={OpenAI Blog},
  volume={1},
  number={8},
  pages={1},
  year={2024}
}

@article{guo2025deepseek,
  title={Deepseek-r1: Incentivizing reasoning capability in llms via reinforcement learning},
  author={Guo, Daya and Yang, Dejian and Zhang, Haowei and Song, Junxiao and Zhang, Ruoyu and Xu, Runxin and Zhu, Qihao and Ma, Shirong and Wang, Peiyi and Bi, Xiao and others},
  journal={arXiv preprint arXiv:2501.12948},
  year={2025}
}

@article{wang2025timecausality,
  title={TimeCausality: Evaluating the Causal Ability in Time Dimension for Vision Language Models},
  author={Wang, Zeqing and Zhang, Shiyuan and Tang, Chengpei and Wang, Keze},
  journal={arXiv preprint arXiv:2505.15435},
  year={2025}
}

@article{komanduri2025causalvlbench,
  title={CausalVLBench: Benchmarking Visual Causal Reasoning in Large Vision-Language Models},
  author={Komanduri, Aneesh and Bhaila, Karuna and Wu, Xintao},
  journal={arXiv preprint arXiv:2506.11034},
  year={2025}
}

@inproceedings{thrush2022winoground,
  title={Winoground: Probing vision and language models for visio-linguistic compositionality},
  author={Thrush, Tristan and Jiang, Ryan and Bartolo, Max and Singh, Amanpreet and Williams, Adina and Kiela, Douwe and Ross, Candace},
  booktitle={Proceedings of the IEEE/CVF Conference on Computer Vision and Pattern Recognition},
  pages={5238--5248},
  year={2022}
}

@article{hua2024mmcomposition,
  title={Mmcomposition: Revisiting the compositionality of pre-trained vision-language models},
  author={Hua, Hang and Tang, Yunlong and Zeng, Ziyun and Cao, Liangliang and Yang, Zhengyuan and He, Hangfeng and Xu, Chenliang and Luo, Jiebo},
  journal={arXiv preprint arXiv:2410.09733},
  year={2024}
}

@inproceedings{du2024uncovering,
  title={Uncovering what why and how: A comprehensive benchmark for causation understanding of video anomaly},
  author={Du, Hang and Zhang, Sicheng and Xie, Binzhu and Nan, Guoshun and Zhang, Jiayang and Xu, Junrui and Liu, Hangyu and Leng, Sicong and Liu, Jiangming and Fan, Hehe and others},
  booktitle={Proceedings of the IEEE/CVF Conference on Computer Vision and Pattern Recognition},
  pages={18793--18803},
  year={2024}
}

@misc{gemini_pro_2024,
  title        = {Gemini Pro},
  author       = {{Google DeepMind}},
  howpublished = {\url{https://deepmind.google/models/gemini/pro/}},
  year         = {2024},
  note         = {Accessed: 2025-11-11}
}

@misc{openai_gpt4o_2024,
  title        = {Hello GPT-4o},
  author       = {{OpenAI}},
  howpublished = {\url{https://openai.com/index/hello-gpt-4o/}},
  year         = {2024},
  note         = {Accessed: 2025-11-11}
}

@article{yang2025qwen3,
  title={Qwen3 technical report},
  author={Yang, An and Li, Anfeng and Yang, Baosong and Zhang, Beichen and Hui, Binyuan and Zheng, Bo and Yu, Bowen and Gao, Chang and Huang, Chengen and Lv, Chenxu and others},
  journal={arXiv preprint arXiv:2505.09388},
  year={2025}
}

@article{zhang2023video,
  title={Video-llama: An instruction-tuned audio-visual language model for video understanding},
  author={Zhang, Hang and Li, Xin and Bing, Lidong},
  journal={arXiv preprint arXiv:2306.02858},
  year={2023}
}

@inproceedings{maaz2024video,
  title={Video-chatgpt: Towards detailed video understanding via large vision and language models},
  author={Maaz, Muhammad and Rasheed, Hanoona and Khan, Salman and Khan, Fahad},
  booktitle={Proceedings of the 62nd Annual Meeting of the Association for Computational Linguistics (Volume 1: Long Papers)},
  pages={12585--12602},
  year={2024}
}

@article{jin2023cladder,
  title={Cladder: Assessing causal reasoning in language models},
  author={Jin, Zhijing and Chen, Yuen and Leeb, Felix and Gresele, Luigi and Kamal, Ojasv and Lyu, Zhiheng and Blin, Kevin and Gonzalez Adauto, Fernando and Kleiman-Weiner, Max and Sachan, Mrinmaya and others},
  journal={Advances in Neural Information Processing Systems},
  volume={36},
  pages={31038--31065},
  year={2023}
}

@article{chen2025counterbench,
  title={Counterbench: A benchmark for counterfactuals reasoning in large language models},
  author={Chen, Yuefei and Singh, Vivek K and Ma, Jing and Tang, Ruxiang},
  journal={arXiv preprint arXiv:2502.11008},
  year={2025}
}

@inproceedings{agrawal2018don,
  title={Don't just assume; look and answer: Overcoming priors for visual question answering},
  author={Agrawal, Aishwarya and Batra, Dhruv and Parikh, Devi and Kembhavi, Aniruddha},
  booktitle={Proceedings of the IEEE conference on computer vision and pattern recognition},
  pages={4971--4980},
  year={2018}
}

@inproceedings{zellers2019recognition,
  title={From recognition to cognition: Visual commonsense reasoning},
  author={Zellers, Rowan and Bisk, Yonatan and Farhadi, Ali and Choi, Yejin},
  booktitle={Proceedings of the IEEE/CVF conference on computer vision and pattern recognition},
  pages={6720--6731},
  year={2019}
}

@article{lake2017building,
  title={Building machines that learn and think like people},
  author={Lake, Brenden M and Ullman, Tomer D and Tenenbaum, Joshua B and Gershman, Samuel J},
  journal={Behavioral and brain sciences},
  volume={40},
  pages={e253},
  year={2017},
  publisher={Cambridge University Press}
}

@article{marcus2020next,
  title={The next decade in AI: four steps towards robust artificial intelligence},
  author={Marcus, Gary},
  journal={arXiv preprint arXiv:2002.06177},
  year={2020}
}

@article{liu2025causal3d,
  title={Causal3d: A comprehensive benchmark for causal learning from visual data},
  author={Liu, Disheng and Qiao, Yiran and Liu, Wuche and Lu, Yiren and Zhou, Yunlai and Liang, Tuo and Yin, Yu and Ma, Jing},
  journal={arXiv preprint arXiv:2503.04852},
  year={2025}
}

@article{yu2024mm,
  title={Mm-vet v2: A challenging benchmark to evaluate large multimodal models for integrated capabilities},
  author={Yu, Weihao and Yang, Zhengyuan and Ren, Lingfeng and Li, Linjie and Wang, Jianfeng and Lin, Kevin and Lin, Chung-Ching and Liu, Zicheng and Wang, Lijuan and Wang, Xinchao},
  journal={arXiv preprint arXiv:2408.00765},
  year={2024}
}

@inproceedings{parcalabescu2022valse,
  title={VALSE: A task-independent benchmark for vision and language models centered on linguistic phenomena},
  author={Parcalabescu, Letitia and Cafagna, Michele and Muradjan, Lilitta and Frank, Anette and Calixto, Iacer and Gatt, Albert},
  booktitle={Proceedings of the 60th Annual Meeting of the Association for Computational Linguistics (Volume 1: Long Papers)},
  pages={8253--8280},
  year={2022}
}

\clearpage

\setcounter{page}{1}

\begin{center}
    {\textbf{SUMMARY OF THE APPENDIX}} \\[12pt]
\end{center}
This appendix provides additional results, implementation details, and analyses. 
It is organized as follows:

\begin{itemize}
    \item \S A describes the environment settings and implementation parameters used across all experiments.
    \item \S B presents the ablation experiments and case studies, including CFGPT component analysis and representative error cases.
    \item \S C provides dataset statistics, covering video distributions, question distributions, and the impact of training data quality.
    \item \S D reports causal graph evaluation metrics, including human-verified causal-edge correctness and quantitative quality assessments.
    \item \S E provides Chain-of-Thought readability metrics, summarizing human evaluation protocols.
\end{itemize}

\vspace{8pt}

\appendix

\section{Environment Setting and Parameters}
\subsection{Dataset Generation}
\paragraph{Multi-Agent Causal Graph Construction}
We employ the latest \textbf{DeepSeek-V3.1 (236B)} as the backbone LLM for all four agents in the multi-agent causal discovery pipeline which is in Section 2.3. Here are the prompts of the LLM Observer, Verifier, Critic, and Synthesizer Agents in figure~\ref{fig:Observer}, \ref{fig:Verifier}, \ref{fig:Critic}, and \ref{fig:Synthetic}. 
\begin{figure}[h!]
  \centering
  \begin{mdframed}[backgroundcolor=white, linewidth=1pt, linecolor=black]
   \textbf{Prompt:} "You are an efficient AI assistant acting as a causal spotter. Your task is to quickly evaluate a list of action pairs from a video and rank them based on their potential for a direct causal relationship. Do not perform a full, deep analysis. Use your common sense, causal knowledge and understanding of physical interactions to provide a quick causal likelihood confidence score. Your entire response MUST be a single, valid JSON object."
  \end{mdframed}
  \caption{Observer Agent Prompt}
  \label{fig:Observer}
\end{figure}

\begin{figure}[h!]
  \centering
  \begin{mdframed}[backgroundcolor=white, linewidth=1pt, linecolor=black]
   \textbf{Prompt:} "You are a Causal Analyst and an expert in the specified domain. Your task is to rigorously evaluate a single proposed causal link between two actions. Perform a Abduction and Counterfactual Test and a Backdoor Criterion Test. Provide your output in a single, valid JSON object. $\langle \text{few\_shot\_1} \rangle$, $\langle \text{few\_shot\_2} \rangle$."
  \end{mdframed}
  \caption{Verifier Agent Prompt}
  \label{fig:Verifier}
\end{figure} 

\begin{figure}[h!]
  \centering
  \begin{mdframed}[backgroundcolor=white, linewidth=1pt, linecolor=black]
   \textbf{Prompt:} "You are a skeptical Adversarial Critic. Your goal is to find flaws in the Verifier's causal analysis. Challenge their reasoning, even if it seems correct. Provide your output in a single, valid JSON object. $\langle \text{few\_shot\_1} \rangle$, $\langle \text{few\_shot\_2} \rangle$."
  \end{mdframed}
  \caption{Critic Agent Prompt}
  \label{fig:Critic}
\end{figure}
\begin{figure}[h!]
  \centering
  \begin{mdframed}[backgroundcolor=white, linewidth=1pt, linecolor=black]
   \textbf{Prompt:} "You are a lead causal analyst acting as the final judge. Weigh the arguments from a Verifier and a Critic to make a final, justified decision. Your entire response MUST be a single, valid JSON object. $\langle \text{Verifier Statements} \rangle$, $\langle \text{Critic Statements} \rangle$."
  \end{mdframed}
  \caption{Synthetic Agent Prompt}
  \label{fig:Synthetic}
\end{figure}
\paragraph{Questions Generation}
After video filtered by metrics, we retain 712 videos that exhibit sufficient causal complexity. We then apply text-based LLMs to synthesize counterfactual questions conditioned on each video's causal graph and associated event descriptions. For every video and each of the three counterfactual levels, the LLM generates ten candidate questions. The full prompt templates used in this process are provided below.

\begin{figure*}[h!]
  \centering
  \begin{mdframed}[backgroundcolor=white, linewidth=1pt, linecolor=black]
   \textbf{Prompt:} You are a senior question designer for a video-based causal reasoning benchmark. \\
    Your input will be:\\
    - An ordered action chain from a single video (\textbf{steps}: each has an \textbf{id}, \textbf{text}, \textbf{timestamp}).\\
    - The causal graph ($\text{causal\_links}$: a list of $[\text{from\_id}$, $\text{to\_id}]$ pairs).\\
    - A list of critical or non-default actions ($\text{key\_actions}$) identified by prior analysis. Use these as inspiration for high-value questions.\\

    Your task:
    Generate \textbf{three categories} of counterfactual questions. Each category should have \textbf{10 candidate questions} (total 30). For every candidate questions, produce: brief answers with short descriptions in English.

    Formatting \& Output:
    Return a single JSON object with the exact schema provided in the user task.
    
    Generation rules (CRUCIAL):\\
    1) \textbf{Q1 Adjacent-not-causal test}:\\
       - Pick \textbf{adjacent} actions A, B from steps that are \textbf{NOT} directly linked in $\text{causal\_links}$. \\
       - Write a counterfactual: “ What if A had not occur, would B still occur?” \\
       - The goal is to test against the "temporal adjacency = causality" fallacy. \\
    
    2) \textbf{Q2 Multi-hop (A→...→E)}:\\
       - Use $\text{causal\_links}$ to find a path of length $> 1$, such as $A \rightarrow C$ via $B$. $A$ and $C$ should not be directly linked. \\
       - The question must only mention A and the endpoint. Ask “If if A had not occur, will $\langle endpoint \rangle$ still occur?” \\
       - Favor using actions from $\text{key\_actions}$ as the intervention point 'A'.
    
    3) \textbf{Q3 Semantic decoy (counterfactual on an non-exist action)}:\\
       - Identify the video’s main goal from the $\text{task\_name}$ and action list.\\
       - Construct a \textbf{plausible, strongly related action that is ABSENT} from the steps.\\
       - Ask: “If the agent did not perform [decoy], would [main goal] still be completed?”\\
       - The answer is often "Yes, it would still succeed" because the decoy, while plausible, was not necessary in this specific video. \\
    
    General constraints: \\
    - Each question must be answerable \textbf{only by watching the video}. \\
    - Keep brief answers concise ($\leq 2$ sentences), bilingual.
  \end{mdframed}
  \caption{Candidate Questions Generation Prompt}
  \label{fig:generation}
\end{figure*}

\subsection{Post-Training}
\paragraph{Video Preprocessing}
In the post-training, All post-training experiments are conducted on \textbf{4$\times$NVIDIA H200 GPUs}. We uniformly sample 512 frames per video using the \textbf{Decord} library. Frames are resized to 144$\times$144 and center-cropped to the vision encoder’s native resolution. We adopt this configuration as it balances temporal completeness with computational feasibility, providing adequate coverage of CounterVQA videos (average duration 199.7s) while remaining compatible with the model’s maximum token budget and keeping inference and training computationally manageable.
\paragraph{Stage I: Cross-Modal Causal Transfer}
In this stage, We perform SFT in Stage I of CFGPT using LoRA on the frozen VLM backbone. we set leanring rate to be $5e-5$, LoRA rank to be $16$. Epoch is $3$.
\paragraph{Stage II: Visual-Causal Alignment Optimization}
We implement GRPO with the following rewards: Causal reward and Visual reward. The weights of each reward are $0.5$. Additionally, Number of samples per question ($K$) is $4$. Learning Rate is $1e-5$. Temperature is $0.9$. Epoch is $3$. 

\section{Ablation Experiment and Case Study}
\subsection{Ablation Experiment on CFGPT Components}
We ablate the two core designs of CFGPT on Qwen-3-VL-8B using the full 3,000+ CounterVQA QA pairs.

\begin{table*}[h]
\centering

\small
\scalebox{0.8}{
\begin{tabular}{l c c c c c}
\toprule
Method & Overall & L1 & L2 & L3 & $\Delta$ \\
\midrule  
Vanilla (zero-shot) & 60.1 & 54.6 & 62.1 & 65.3 & -12.5 \\
\quad + Standard SFT + GRPO (w/o Causal Graph Reward)& 66.3 & 63.4 & 67.6 & 68.6 & -6.3 \\
\quad + SFT + GRPO (w/o Videos Distillation)& 65.9 & 63.9 & 65.7 & 68.0 & -6.7 \\
\quad + GRPO (w/o SFT and Distillation) & 63.7 & 58.3 & 65.7 & 68.0 & -8.9 \\
\quad + Full CFGPT (ours, with Causal Graph Reward and Cross-modal distillation) & \textbf{72.6} & \textbf{70.1} & \textbf{71.6} & \textbf{76.0} & 0.0 \\
  
\bottomrule
\end{tabular}}

\caption{Ablation of Core Component of Stage I and Stage II}
\label{table:ablation}
\end{table*}

As shown in Table~\ref{table:ablation}, each design choice in CFGPT contributes substantially to the final performance. 
Compared to the vanilla Qwen-3-VL 8B baseline (60.1 overall), our full model reaches 72.6 overall accuracy and 70.1,71.6,76.0 on Levels 1 to 3, 
corresponding to gains of +12.5 points overall and up to +10.7 points on the Level-3 questions.

To disentangle where these gains come from, we examine three ablated variants, each removing a key component while keeping the rest of the pipeline unchanged. 
Replacing our structured \textbf{Causal Graph Reward} with a standard binary correctness reward (\emph{Standard SFT + GRPO}) reduces the overall accuracy to 66.3 (–6.3 points relative to Full CFGPT), 
with the largest drop on Level 3 (from 76.0 to 68.6). 
This indicates that coarse 0/1 supervision is insufficient to guide the model toward fine-grained causal structures, especially for non-existent-event counterfactuals.

When we keep the causal-graph-based reward but remove the \textbf{video distillation} component in Stage I (\emph{SFT + GRPO w/o Videos Distillation}), 
performance similarly decreases to 65.9 (–6.7 overall). 
The degradation is spread across all three levels, suggesting that cross-modal distillation from a stronger text teacher is important for aligning the video encoder with temporally structured, language-like causal patterns.

Finally, using \textbf{GRPO alone} without any SFT or distillation (\emph{GRPO w/o SFT and Distillation}) yields 63.7 overall, which is slightly better than the vanilla model but clearly worse than any SFT-based variant (–8.9 compared to Full CFGPT). 
This shows that reinforcement learning by itself can provide some refinement but cannot bootstrap robust counterfactual reasoning without an initial supervised causal prior.

Taken together, these ablations demonstrate that CFGPT benefits from both components: 
(1) the Causal Graph Reward, which enforces structurally consistent counterfactual reasoning, and 
(2) cross-modal distillation in Stage I, which narrows the visual–textual reasoning gap by transferring strong textual causal reasoning into the video backbone.






\subsection{Error Analysis}
\begin{figure}[htbp]
  \centering
  \includegraphics[width=0.45\textwidth]{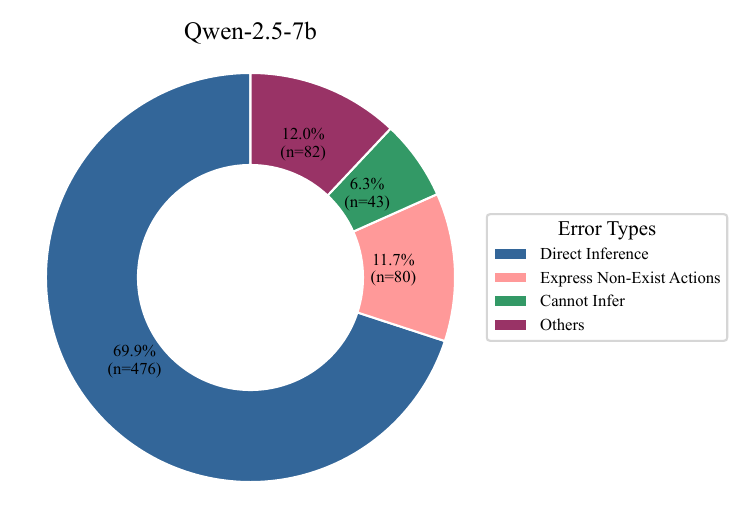 }
  \caption{Error analysis of Qwen-2.5-7B on CounterVQA. Although most predictions come from direct inference without combining with vision (69.9\%), a notable portion involves nonexistent-action reasoning (11.7\%), cannot infer based on inputs errors (6.3\%), or other error modes (12.0\%).}
  \label{fig:error_distribution}
\end{figure}

In order to find out the reason about Qwen-2.5-7b performs below 10 percent accuracy. We have an error analysis and conduct an experiment for perception. The experiment precedure is we add an question about detection result, such as the original question is "If the oven had not been preheated, would the omelet still be completed?" and The posterior question is added as "Does Preheated oven event happen?". The result shows that Qwen-2.5-7b can answer 98.2\% questions about detection questions although it does not improve the reasoning questions result. The error analysis result is as follow. Despite its strong perceptual ability, the distribution of errors shows that the model’s errors arise mainly from reasoning rather than recognition. As illustrated in Figure~\ref{fig:error_distribution}, only 69.9\% of predictions correspond to correct direct inference, while the remaining cases fall into several distinct failure categories. A notable portion of errors (11.7\%) involves reasoning about actions or events that do not exist in the video, indicating incorrect counterfactual grounding. Another 6.3\% of cases reflect situations where the model cannot infer an answer even when all necessary visual evidence is present, revealing a weakness in connecting perceptual observations to causal conclusions. The remaining 12.0\% belong to miscellaneous inconsistencies. Overall, these results suggest that the performance drop is driven by limitations in structured causal reasoning rather than deficiencies in visual understanding.

\begin{figure*}[htbp]
  \centering
  \includegraphics[width=0.95 \textwidth]{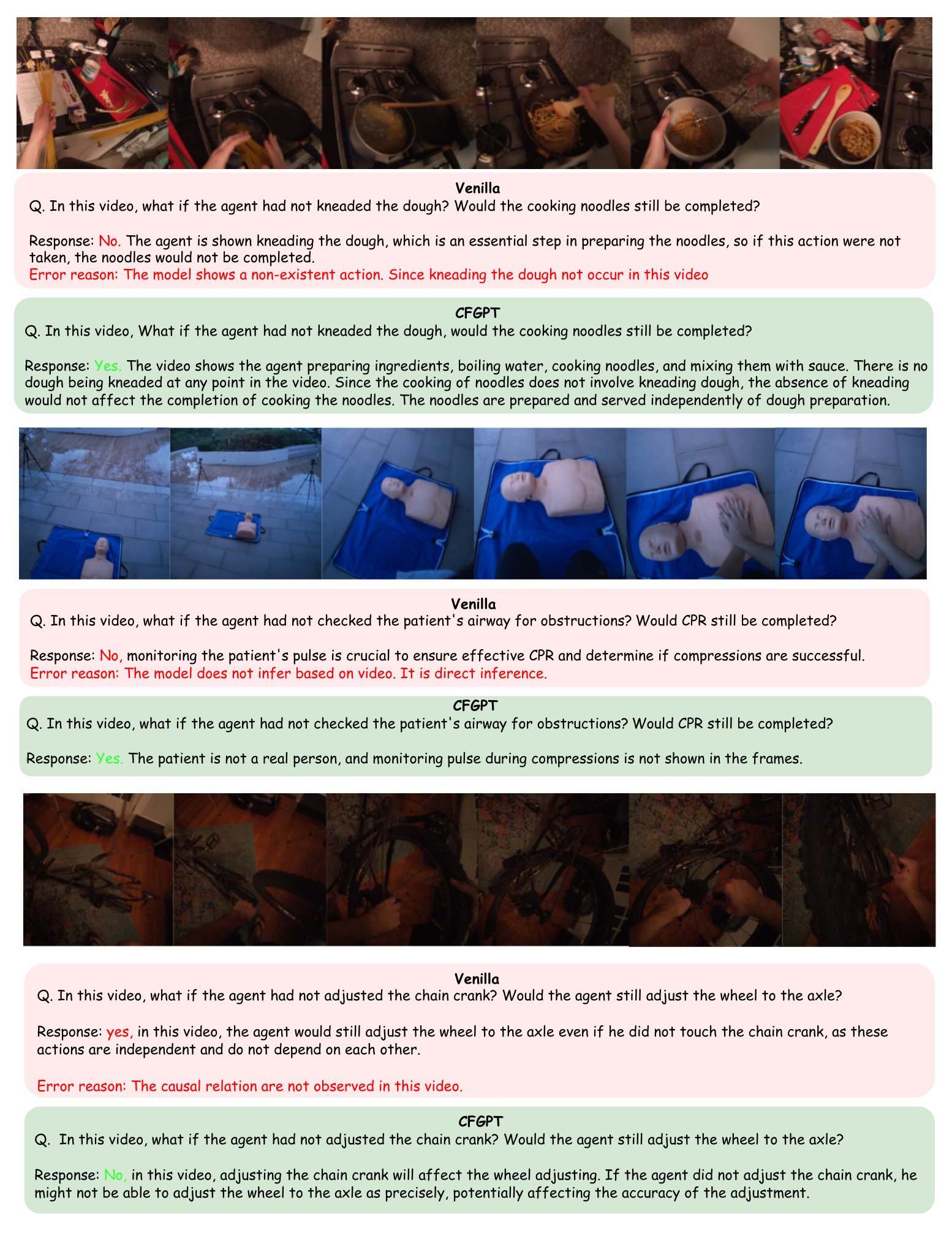}
  \caption{Representative error cases comparing Vanilla and CFGPT models. CFGPT corrects non-existent-action errors, avoids unsupported direct inference, and better captures causal dependencies after finetuning.}
  \label{fig:error_cases1}
\end{figure*}

Figure~\ref{fig:error_cases1} shows three representative failure cases from the vanilla model and the corresponding corrections made by CFGPT. In the first example, the vanilla model hallucinates a non-existent action (“kneading the dough”), while CFGPT correctly identifies that no such action appears in the video and produces a valid counterfactual answer. In the second example, the vanilla model answers purely based on prior knowledge rather than visual evidence, whereas CFGPT grounds its response in what is actually shown. In the third example, the vanilla model misses the causal dependency between adjusting the chain crank and adjusting the wheel, while CFGPT correctly identifies the causal link and provides a consistent counterfactual prediction.

\section{Dataset Statistics}
\subsection{Video Statistics}
\begin{figure}[ht]
  \centering
  \includegraphics[width=0.45\textwidth]{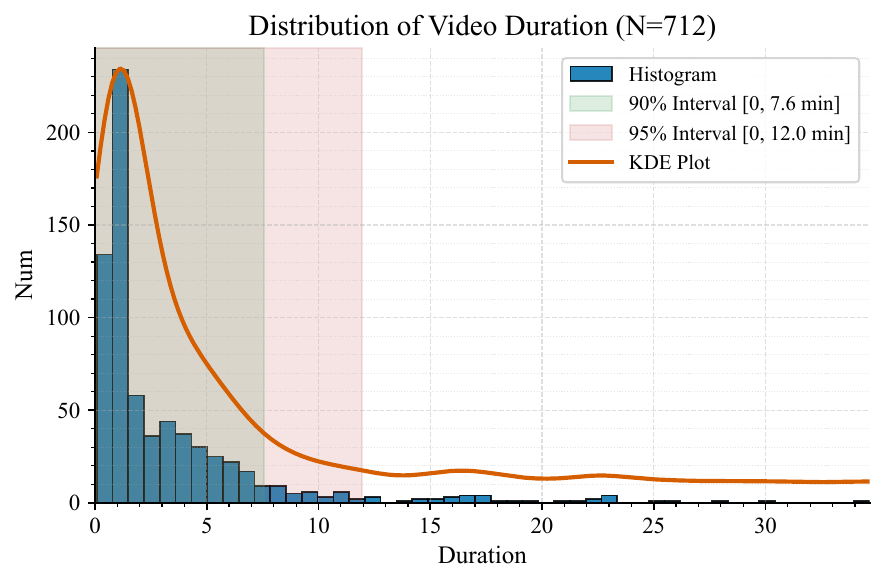}
  \caption{Distribution of video durations in CounterVQA (N=712). 
The histogram and KDE curve show that most videos are short, 
with 90\% of durations below 7.6 minutes and 95\% below 12.0 minutes.}
  \label{fig:video_duration_distribution}
\end{figure}

We use 1200 videos and filter a lot of them. The rest of videos are over 700 videos. 39.3\% of the videos are Human-to-Human (H2H) interactions involving collaborative activities, such as basketball, soccer, or dance. 60.7\% are Human-to-Object (H2O) interactions focus on tasks where a person manipulates objects, such as cooking, climbing, or COVID-19 testing. It requires understanding of physical causality and how actions affect material states. We analyze the temporal characteristics of the 712 selected videos in CounterVQA. As shown in Figure~\ref{fig:video_duration_distribution}, the duration distribution is heavily skewed toward short videos: approximately 90\% of the samples are under 7.6 minutes, and 95\% are under 12.0 minutes. This indicates that most videos concentrate within a narrow temporal range of a few minutes. These statistics justify our temporal sampling strategy, which aims to provide sufficient temporal coverage while remaining compatible with the model’s maximum input length and ensuring efficient training and inference performance.

\subsection{Questions Statistics}
Our final question set contains 3987 counterfactual QA pairs.  
Roughly 36.2\% of the questions are derived from H2H videos, while 63.8\% 
originate from H2O videos, indicating that the dataset naturally leans 
toward object-centric interactions.

In terms of counterfactual complexity, Q1 (adjacent) questions constitute 
approximately 38.6\% of the dataset, Q2 (long-chain) questions 26.7\%, and 
Q3 (non-existent-event) questions 34.6\%. This distribution maintains a 
balanced coverage across the three reasoning categories, avoiding 
over-concentration in any single type.

\subsection{Impact of Training Data Quality}
Generating high-quality counterfactual questions is inherently challenging. Unlike caption-based question generation, counterfactual queries must be grounded in the actual events of the video, respect causal dependencies, and remain logically valid under hypothetical interventions. Naïvely prompting an LLM with annotations often produces questions that are temporally incorrect, causally ambiguous, or based on events that never occurred.

To better understand the quality differences between question generation approaches, we conducted a human evaluation study with two trained annotators who independently assessed random 120 naïvely generated questions and 120 questions from our multi-agent causal-graph pipeline. Each question was evaluated along two key dimensions: reasoning difficulty which is rated on a 1–5 scale and grounding accuracy. The evaluation revealed substantial improvements from our proposed approach. Questions generated through the multi-agent pipeline consistently demanded deeper causal reasoning, with an average difficulty rating of 3.8 compared to just 2.6 for naïve questions. More importantly, we assessed grounding accuracy which is to assess questions correctly reference events that actually occur in the video with proper temporal ordering. Our pipeline achieved 97\% grounding accuracy compared to only 73\% for naïve generation. This substantial improvement is closely tied to logical validity. Because naïve questions frequently suffer from causal reasoning flaws, hallucinating non-existent events or misrepresenting temporal sequences, whereas our causal-graph-guided approach ensures questions are anchored to verified causal structures. These results demonstrate that explicit causal graph guidance is crucial for generating counterfactual questions that are both intellectually challenging and factually grounded in video content.

\section{Causal Graphs}

To assess the quality of the causal relations produced by our multi-agent system, we conducted human evaluation on a sampled subset of generated causal graphs. Annotators judged whether each predicted causal edge was valid according to the video content and temporal dependencies. As shown in Table~\ref{table:causal_metrics}, the system achieves a precision of 0.8839, a recall of 0.9629, an F1-score of 0.9112, and an overall accuracy of 0.8586. These results indicate that the multi-agent pipeline is able to recover most true causal relations (high recall) while keeping false positives relatively low (high precision).

\begin{table}[htbp]
\centering
\scalebox{0.9}{
\begin{tabular}{lccc}
\toprule
\textbf{Precision} & \textbf{Recall} & \textbf{F1-Score} & \textbf{Accuracy}\\
\midrule
0.8839   &  0.9629  &  0.9112 & 0.8586\\
\bottomrule
\end{tabular}}
\caption{Human-evaluation metrics for the generated causal relations. The multi-agent system achieves high precision (0.8839), recall (0.9629), F1-score (0.9112), and accuracy (0.8586), indicating strong causal-edge quality and low error rates.}
\label{table:causal_metrics}
\end{table}
The remaining errors typically arise from visually subtle interactions or cases where temporal adjacency does not correspond to true causality, highlighting the intrinsic difficulty of causal graph construction in multi-step activities. This difficulty further motivates our rigorous question filtering stage: only questions that require non-trivial causal reasoning and cannot be solved by simpler vision-language models are retained. This ensures that the final dataset contains challenging and causally meaningful supervision signals.

\section{CoT Readability Metrics}
To assess the clarity and coherence of model-generated chain-of-thought (CoT) explanations, we conduct a human-evaluation study and define a Readability Metric scored on a 0–5 scale. Evaluators are asked to judge the step-by-step reasoning only (not correctness of the final answer), focusing on linguistic quality and logical structure. Each type of CoT output is evaluated using 100 randomly sampled instances, independently scored by two annotators with professional backgrounds in computer vision and causal inference. The final readability score is computed as the average of the two ratings.

\begin{itemize}
    \item \textbf{5 -- Excellent:} Reasoning is fluent, logically coherent, well-structured, and easy to follow; no ambiguous or redundant expressions.
    \item \textbf{4 -- Good:} Mostly clear and well-organized; minor linguistic issues but reasoning remains easy to understand.
    \item \textbf{3 -- Fair:} Reasoning is understandable but contains redundant steps, occasional unclear phrasing, or minor inconsistencies.
    \item \textbf{2 -- Poor:} Reasoning is difficult to follow due to disorganized structure or unclear transitions; readability is significantly affected.
    \item \textbf{1 -- Very Poor:} Highly fragmented or repetitive reasoning; logical progression is hard to interpret.
    \item \textbf{0 -- Unreadable:} Output is incoherent, severely ungrammatical, or does not constitute meaningful reasoning.
\end{itemize}


\end{document}